\documentstyle[eaclap]{article}

\title{\vspace{-0.5in}The Semantics of Motion}
\author{Pierre Sablayrolles \\
      I R I T --- Universit\'e Paul Sabatier \\
      118 route de Narbonne \\
      31062 Toulouse - France \\
      phone  : +33 61 55 67 64 \\
      fax    : +33 61 55 83 25 \\
      e-mail : sablay@irit.fr \\}

\begin{document}

\maketitle

\begin{abstract}
   In this paper we present a semantic study of motion complexes (ie. of
   a motion verb followed by a spatial preposition). We focus on the
   spatial and the temporal intrinsic semantic properties of the motion
   verbs, on the one hand, and of the spatial prepositions, on the other
   hand. Then we address the problem of combining these basic semantics
   in order to formally and automatically derive the spatiotemporal
   semantics of a motion complex from the spatiotemporal properties
   of its components.
\end{abstract}

\section{Introduction}

Most of natural languages provide two types of lexical items to describe
the motion of an entity with respect to some location: {\bf motion verbs}
({\em to run}; {\em to enter}) and {\bf spatial prepositions} ({\em from};
{\em towards}). Motion verbs can be used directly with a location, when they
are transitive ({\em to cross the town}) or with a spatial preposition, when
they are intransitive ({\em to go through the town}). The latter case is more
interesting: most of the French motion verbs are intransitive and the
interaction between motion verbs and spatial prepositions gives detailed
informations about the way human beeings mentally represent spatiotemporal
aspects of a motion. When we describe a motion, the fact to choose a
verb instead of another, a preposition instead of another, a syntactic
structure instead of another, reveals our mental cognitive representation.
We claim that natural languages can be considered as a trace of these
representations, in
which it is possible, with systematic and detailled linguistic studies, to
light up the way spatiotemporal properties are represented and on which
basic concepts
these representations lie. We present such linguistic investigations on French
motion verbs and spatial prepositions and the basic concepts we have found.
We also address compositional semantics for motion complexes (ie. a motion
verb followed by a spatial preposition) and show that the complexity and the
refinements of the linguistic studies presented just before are justified
and required at the compositional level in order to capture the different
behaviours in the compositional processes that exist with the French language.
We also compare with the English language and draw some conclusions on the
benefits of our approach.

\section{Lexical Semantics for Motion Verbs}

Following Gruber (1965), Jackendoff (1976), Boons (1985), we approach
motion verbs in terms of some ``localist semantical'' role labels.
The linguistic study of French intransitive motion verbs (see eg. (Asher \&
Sablayrolles, 1994a)) we have realized has allowed the
definition of an ontology for ``location'' in three basic concepts:

\begin{itemize}
   \item {\bf locations} which are concrete places
         ({\em a room; a house; a street});
   \item {\bf positions} which are parts of a location
         ({\em the position where I am in this room});
   \item {\bf postures} which are ways to be in a position
         ({\em to be standing, sitting, lying}).
\end{itemize}

With the help of this ontology we have
realized a typology for intransitive motion verbs. We distinguish 4 categories
on the basis of which kind of ``location'' they intrinsically refer to.

\begin{itemize}
   \item {\bf Change of location (CoL)} verbs ({\tt entrer}--{\em to enter};
         {\tt sortir}--{\em to go out}) denote a change of location. When we
         enter some place or go out of some place, we have different spatial
         relation with the location (ie. inside/outside) before and after the
         motion.
   \item {\bf Change of position} verbs ({\tt voyager}--{\em to travel};
         {\tt courir}--{\em to run}) denote a change of position. When we
         travel or run, we go from some part to another part of a same
         global location. Such verbs do not behave all homogeneously.
         \begin{itemize}
            \item Some denote a change of position which always occur
                  ({\tt voyager}--{\em to travel}). For example, we cannot say
                  {\tt voyager sur place}--{\em to travel in place}. We call
                  these verbs {\bf change of position (CoPs)} verbs.
            \item Others denote only possible change of position
                  ({\tt courir}--{\em to run}). For example, we can say
                  {\tt courir sur place}--{\em to run in place}. We call
                  these verbs {\bf inertial change of position (ICoPs)} verbs.
         \end{itemize}
   \item {\bf Change of posture (CoPtu)} verbs
         ({\tt s'asseoir}--{\rm to sit down};
         {\tt se baisser}--{\em to bend down}).
         They denote a change of the relations between the parts of an entity.
\end{itemize}

For the following, we will focus on {\bf CoL} verbs (the Change of Location
verbs), mainly because they are rich in spatiotemporal informations, but also
because we have at disposal exhaustive lists of French CoL verbs.
We have realized a systematic and fine linguistic study on these verbs,
looking carefully at each of them, one by one (and we have 440 CoL verbs in
French), in order to extract their intrinsic spatiotemporal properties.
These semantic properties can be characterized by a restructuration of the
space induced by the so-called {\bf reference location (lref)}
(cf. (Talmy, 1983)). This lref, implicitly suggested by each CoL verb,
can be either the initial location (as with {\tt partir}--{\em to leave};
{\tt sortir}--{\em to go out}), or the path
({\tt passer, traverser}--{\em to pass through}) or the final location
({\tt arriver}--{\em to arrive};
{\tt entrer}--{\em to enter}) of the motion. Indeed, verbs like
{\tt sortir} intrinsically suggest a location of which we have gone out.
This space, induced by the lref, is characterized by most of the authors in
the literature by a two-part spatial system consisting in
the inside and the outside of the lref. We propose to refine this structure
with two new concepts, required to distinguish minimal pairs like {\tt sortir}
({\em to go out})/{\tt partir} ({\em to leave}), and {\tt entrer}
({\em to enter})/{\tt atterir} ({\em to land}). These concepts are:

\begin{enumerate}
   \item a limit of proximity distinguishing an outside of proximity from a
         far away outside; indeed, if {\tt sortir} simply requires to go out
         of the lref, {\tt partir} in addition forces the mobile to go
         sufficiently far away from that lref.
   \item an external zone of contact required by verbs like {\tt atterir} for
         which the final location is neither the lref (in contrast with
         {\tt entrer}) or the outside (or proximity outside) of the lref
         (in contrast with {\tt s'approcher}--{\em to approach}).
\end{enumerate}

We have so defined a structuration of the space based on 4 zones :

\begin{itemize}
   \item the {\bf inside};
   \item the {\bf external zone of contact};
   \item the {\bf outside of proximity};
   \item the {\bf far away outside}.
\end{itemize}

This structuration is close to the
way Jackendoff and Landau (1992) encode the space induced by the reference
object introduced by a static spatial preposition. As we have come to these
distinctions by examining different linguistic material, we conclude that
language structures space in the same way whatever sort of lexical items
(motion verbs (dynamic)/(static) spatial prepositions) we examine.
This has allowed us to classify CoL verbs into 10 classes on the basis of
which zones the mobile is inside, at the beginning and at the end of its
motion. Note that all the geometrical possibilities are not lexicalized
in French.

\section{Lexical Semantics for Spatial Prepositions}

We have followed the same approach with French spatial prepositions, but
using a structuration of the space induced by the location introduced in
the PP by the preposition, and not induced by the lref as for verbs.
Following Laur (1993), we consider simple prepositions (like {\em in}) as
well as prepositional phrases (like {\em in front of}). We have classified
199 such French prepositions into 16 groups using in addition of our zones
two other criteria:

\begin{itemize}
  \item prepositions can be:
        \begin{itemize}
           \item positional (like {\em in})
           \item directional (like {\em into})
        \end{itemize}
  \item directional prepositions can be:
        \begin{itemize}
           \item Initial (like {\em from})
           \item Medial (like {\em through})
           \item Final (like {\em to})
        \end{itemize}
        depending if they focus on the initial location, on the path or on
        the final location of the motion.
\end{itemize}

\section{Compositional Semantics for Motion Complexes}

The linguistic studies, used for the typologies of CoL verbs and spatial
prepositions, have been realized on verbs considered without any adjuncts,
in their atemporal form and independently of any context, on the one hand,
and on prepositions considered independently of any context, on the other.
This methodology, discussed in Borillo \& Sablayrolles (1993), has allowed
us to extract the intrinsic semantics of these lexical items.

Since natural languages put together verbs and prepositions in a sentence,
we have developped a formal calculus (see (Asher \& Sablayrolles, 1994b)),
based on these two typologies, which computes, in a compositional way, the
spatiotemporal properties of a motion complex from the semantic properties
of the verb and of the preposition. For reason of space we cannot detail our
formalism here, but we intend to present it in the talk.

The semantics of a motion complex is not the simple addition of the semantics
of its constituents. On the contrary, it is the result of a complex
interaction between these properties. It is often the case that from this
interaction appear new properties that belong neither to the verb or the
preposition. These new properties are only the result of the interaction of
the verb with the preposition.

Let us consider for example the following VP:
{\tt sortir dans le jardin}--{\em to go out into the garden}.
The verb {\tt sortir}--{\em to go out} implicitly suggests an initial
location; the preposition {\tt dans}--(which means {\em in},
but which is translated here by {\em into}) is a
positional preposition and, as so, only denotes the static spatial
relation {\it inside}. The location {\tt le jardin}--{\em the garden}
is the final location of the motion. This {\it final} information was
contained neither in the verb or in the preposition. This information is the
result of the interaction of the verb {\tt sortir}--{\em to go out} with
the preposition {\tt dans}--{\em in/into}.

Note that the combination for such items does not behave the same in
English, where the {\it final} information is explicitly brought by the
preposition {\em into}, which is a directional preposition, and where this
particular combination does not create new information.

This shows the neccesity to take into account such language specific behaviour
in natural languages understanding systems and in natural languages machine
translation. We formalize with 11 axioms in a non-monotonic first order logic
the behaviour of all possible kinds of verb-preposition association for the
French language. We use non-monotonic logic in order to represent defeasible
or generic rules and also in order to encode defaults about lexical entries.

These axioms are based on the lexical semantics of CoL verbs and of spatial
prepositions. They also take into account the syntactic structure of the
sentence (we have supposed an X-bar syntax with a VP internal subject, though
this is not essential) and the links which exist at the level of discours
between this sentence and the previous and following sentences of the text.
These links, called discourse relations, are basic concepts on which texts
are structured (cf. (Asher, 1993)).

\section{Conclusion}

The study and the first results we have here presented cover from lexical
semantics to discourse structures, with strong interactions between these two
ends. Indeed, lexical informations can be used to disambiguate the structure
of the discours, as well as discourse relations can be used to disambiguate
lexical entries, as shown in (Asher \& Sablayrolles, 1994b). Our work is
based on systematic and
very detailed linguistic studies which lead to rather complex computations
for calculating the spatiotemporal semantics of a motion complex. But we
have seen that
this level of detail and complexity is necessary if one want to understand,
to formalize and to compute a right spatiotemporal semantics for motion
complexes.
We continue our investigations on two directions:

\begin{enumerate}
   \item we compare our results with similar works in course of realization
         on the Basquian language (by Michel Aurnague) and on the Japanese
         language (by Junichi Saito);
   \item we use the results presented here for refining the notions of the
         Aktionsart, where the structuration of the space in 4 zones can be
         used to distinguish sub-classes inside the traditional well known
         classes of aspectual studies.
\end{enumerate}

\section*{References}

{\bf Nicholas Asher and Pierre Sablayrolles.} 1994a.
   A Compositional Spatio-temporal Semantics for French Motion Verbs and
      Spatial PPs.
   Proccedings of {\em SALT4, Semantics and Linguistic Theory},
      Rochester, NY, May 6-8, 1994.

{\bf Nicholas Asher and Pierre Sablayrolles.} 1994b.
   A Typology and Discourse Semantics for Motion Verbs and Spatial PPs
      in French.
   {\em Journal of Semantics}, in press, 1994.

{\bf Nicholas Asher.} 1993.
   Reference to Abstract Objects in Discourse.
   {\em Kluwer Academic Publishers}, 1993.

{\bf Jean Paul Boons.} 1985.
   Pr\'eliminaires \`a la classification des verbes locatifs : les
      compl\'ements de lieu, leurs crit\`eres, leurs valeurs aspectuelles.
   {\em Linguisticae Investigationes}, 9(2):195-267, 1985.

{\bf Mario Borillo and Pierre Sablayrolles.} 1993.
   The Semantics of Motion Verbs in French.
   Proceedings of the {\em 13th International Conference on Natural Language
      Processing of Avignon}, May 24-28, 1993, Avignon, France.

{\bf J.S. Gruber.} 1965.
   Studies in Lexical Relations.
   {\em Doctoral Dissertation}, MIT, 1965.

{\bf Ray Jackendoff.} 1976.
   Towards an Explanatory Semantic Representation.
   {\em Linguistic Inquiry}, 7:89-150.

{\bf Ray Jackendoff and Barbara Landau.} 1992.
   ``What'' and ``Where'' in Spatial Language and Spatial Cognition.
   {\em BBS report}, Cambridge University Press, 1992.

{\bf Dany Laur.} 1993.
   La relation entre le verbe et la pr\'eposition dans la s\'emantique
      du d\'eplacement.
   {\em Language, La couleur des pr\'epositions}:47-67, June 1993.

{\bf Leonard Talmy.} 1983.
   How Language Structures Space.
   {\em Spatial Orientation: theory, research and application}, Pick and
      Acredolo (eds), Plenum pub. corporation, NY, 1983.

\end{document}